\documentclass{article}

\PassOptionsToPackage{numbers, compress}{natbib}
\usepackage[preprint]{neurips_2025}

\usepackage[utf8]{inputenc} % allow utf-8 input
\usepackage[T1]{fontenc}    % use 8-bit T1 fonts
\usepackage{hyperref}       % hyperlinks
\usepackage{url}            % simple URL typesetting
\usepackage{booktabs}       % professional-quality tables
\usepackage{amsfonts}       % blackboard math symbols
\usepackage{nicefrac}       % compact symbols for 1/2, etc.
\usepackage{microtype}      % microtypography
\usepackage{xcolor}         % colors
\usepackage{float}

\usepackage{lmodern}
\usepackage{amsmath}
\usepackage{tikz}
\usetikzlibrary{positioning,arrows.meta,calc,fit,backgrounds}
\usepackage{adjustbox}
\usepackage{graphicx}

\usepackage{bbm}

\title{Symbolic-Diffusion: Deep Learning Based Symbolic Regression with D3PM Discrete Token Diffusion}

\author{
  Ryan T. Tymkow\\
  Department of Mechatronics Engineering\\
  University of Waterloo\\
  Waterloo, ON Canada, N2L 3G1 \\
  \texttt{rtymkow@uwaterloo.ca} \\
  \And
  Benjamin D. Schnapp \\
  Department of Mechatronics Engineering\\
  University of Waterloo\\
  Waterloo, ON Canada, N2L 3G1 \\
  \texttt{bdschnapp@uwaterloo.ca} \\
 \AND
  Mojtaba Valipour \\
  Department of Computer Science \\
  University of Waterloo \\
  Waterloo, ON Canada, N2L 3G1 \\
  \texttt{mojtaba.valipour@uwaterloo.ca} \\
  \And
  Ali Ghodshi \\
  Department of Statistics and Actuarial Science \\
  University of Waterloo \\
  Waterloo, ON Canada, N2L 3G1 \\
  \texttt{ali.ghodsi@uwaterloo.ca} \\
}

\begin{document}

\maketitle

\begin{abstract}
Symbolic regression refers to the task of finding a closed-form mathematical expression to fit a set of data points. Genetic programming based techniques are the most common algorithms used to tackle this problem, but recently, neural-network based approaches have gained popularity. Most of the leading neural-network based models used for symbolic regression utilize transformer-based autoregressive models to generate an equation conditioned on encoded input points. However, autoregressive generation is limited to generating tokens left-to-right,  and future generated tokens are conditioned only on previously generated tokens. Motivated by the desire to generate all tokens simultaneously to produce improved closed-form equations, we propose Symbolic Diffusion, a D3PM based discrete state-space diffusion model which simultaneously generates all tokens of the equation at once using discrete token diffusion. Using the bivariate dataset developed for SymbolicGPT, we compared our diffusion-based generation approach to an autoregressive model based on SymbolicGPT, using equivalent encoder and transformer architectures. We demonstrate that our novel approach of using diffusion-based generation for symbolic regression can offer comparable and, by some metrics, improved performance over autoregressive generation in models using similar underlying architectures, opening new research opportunities in neural-network based symbolic regression.
\end{abstract}

\section{Introduction}

Symbolic regression refers to the task of finding a closed-form mathematical expression to fit a set of data points, without fixing the mathematical structure of the equation in advance \citep{radwan_comparison_2024}.  Traditional algorithms for this task have primarily utilized genetic-programming based methods to solve this problem, where, inspired by biological evolution, populations of candidate expressions evolve, crossover, and mutate towards maximizing a fitness function \citep{koza_genetic_1994}. While genetic-programming based methods have proven flexible and capable at the task of symbolic regression, and remain the state-of-the-art in terms of the quality of the generated equations \citep{cava_contemporary_2021}, they often suffer from being computationally expensive and are typically slow at inference time as it is necessary to search the entire mathematical space to find the right equation for each dataset \citep{kamienny_end--end_2022}.

Deep-learning-based approaches have received attention in recent research as an alternative to genetic-programming for symbolic regression.  While the performance of deep-learning-based symbolic regression has not been able to match the quality of state-of-the-art genetic-programming-based approaches yet, these methods have excelled at providing decently high quality outputs with significantly faster, near real-time inference \citep{kamienny_end--end_2022}.

Conventionally, the majority of deep learning methods have used autoregressive generation conditioned on encoded input points to produce a tokenized output form of the equation \citep{ kamienny_end--end_2022, valipour_symbolicgpt_2021, vastl_symformer_2022}.  

Diffusion models, which function by learning the reverse process to generate data from noise \citep{sohl-dickstein_deep_2015}, have shown exceptional performance in generative tasks on continuous domains such as image generation \citep{rombach_high-resolution_2022}.  Recently, discrete-space diffusion models have been developed which attain reasonably strong performance in discrete domains such as text generation \citep{austin_structured_2023, lou_discrete_2024, gong_diffuseq_2023}, and has achieved state-of-the-art performance in domains such as music generation \citep{plasser_discrete_2023} or DNA sequencing \citep{li_discdiff_2024}.  The objective of this work was to be the first to develop the a novel symbolic regression model based on discrete token diffusion, and to evaluate the performance of discrete token diffusion for the domain of symbolic regression compared to autoregressive generation.

\subsection{Motivation}
While autoregressive models have demonstrated strong performance across many domains, autoregressive generation remains limited by the sequential generation process, as each future token generated is only conditioned on previously generated tokens, and lacks global context which is needed in the context of Symbolic Regression. 

Discrete state-space diffusion offers a different approach, iteratively denoising the entire sequence simultaneously, allowing the entire sequence to have global context throughout the generation process.  By leveraging the global context available during discrete token diffusion, we hypothesized it may be possible to attain more accurate and syntactically valid symbolic expressions versus autoregressive generation models. 

\subsection{Contributions}
\begin{figure}[h!]
  \centering
  \fbox{%
    \includegraphics[width=0.95\linewidth]{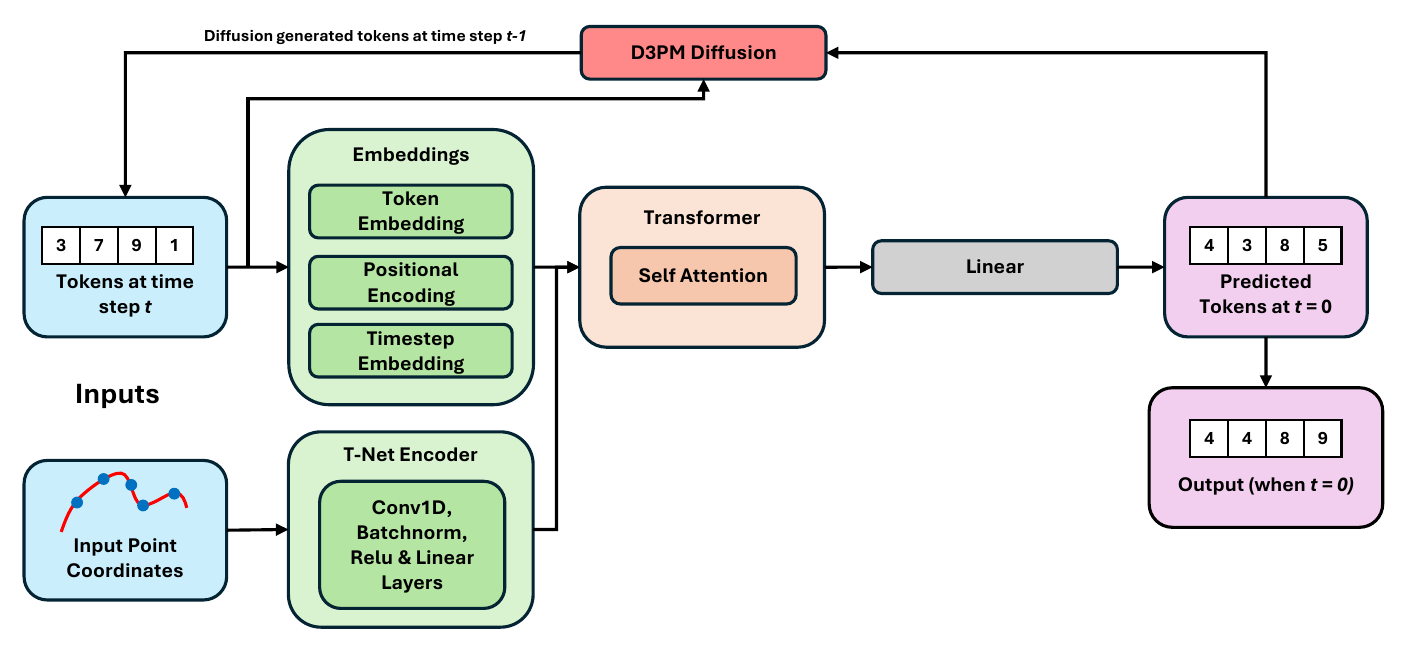} % or .png, adjust width as needed
  }
  \caption{Symbolic Diffusion Architecture}
  \label{fig:d3pm_diffusion}
\end{figure}

In this paper, we present Symbolic Diffusion (Figure \ref{fig:d3pm_diffusion}), which is a symbolic regression model that uses diffusion rather than autoregressive token generation to predict the symbolic form of an equation. This model enables global context and simultaneous generation of all tokens during symbolic regression compared to previous autoregressive approaches.

Symbolic Diffusion utilizes a Pointnet-style feature encoder \citep{qi_pointnet_2017}, followed by a transformer-based decoder designed to predict the ground truth token through a series of denoising steps utilizing the D3PM \citep{austin_structured_2023} discrete state space diffusion architecture .  

Symbolic Diffusion was compared to an autoregressive model based on SymbolicGPT \citep{valipour_symbolicgpt_2021} featuring an identical encoder and transformer architecture, but conditioned to produce tokens sequentially in an autoregressive manner as a GPT model \citep{noauthor_httpscdnopenaicomresearch-coverslanguage-unsupervisedlanguage_understanding_paperpdf_nodate}, versus our models simultaneous diffusion-based generation of all tokens through denoising.  It was found that Symbolic Diffusion performed comparably to the autoregressive baseline, and in some metrics such as mean $R^2$ value, attained a statistically significantly higher value.

Additionally, we provide an open-source implementation of our model, available at \href{https://github.com/rtymk/Symbolic_Diffusion}{https://github.com/rtymk/Symbolic\textunderscore Diffusion.}

\subsection{Related work}
Early work on symbolic regression was based on the concepts of genetic programming, inspired by biological evolution.  Koza (1994) \citep{koza_genetic_1994} applied Darwinian principles of evolution to computational algorithms, with natural selection, mutations, and reproduction with crossover adapting symbolic equations to produce one which best fits a given set of data.  This sparked the birth of classical approaches to symbolic regression.

Genetic algorithms experienced several significant advances over the years improving their performance, such as offspring selection and age layered population structures \citep{burlacu_population_2024} which helped increase the “genetic diversity” of generated equations, improving the performance. 

More recently, Generative Pre Training (GPT) models \citep{radford_improving_nodate} were applied to the domain of symbolic regression in the paper SymbolicGPT by Valipour et al. \citep{valipour_symbolicgpt_2021}.  Autoregressive generation utilizing the transformer architecture \citep{vaswani_attention_2023} was used to generate symbolic equations conditioned on input coordinate data. T-Net embeddings \citep{qi_pointnet_2017} were used to embed the spatial data from the input coordinates.  SymbolicGPT was shown to perform favorably compared to some genetic programming approaches in both performance and accuracy. 

State of the art neural network (Symformer \citep{vastl_symformer_2022}, E2E Symbolic Regression \citep{kamienny_end--end_2022}) use autoregression with end-to-end approaches where constants are predicted to pre-initalize classical constant fitting algorithms, unlike SymbolicGPT \citep{valipour_symbolicgpt_2021} where the model does not predict constants prior to using classical constant fitting.  (Our model was not compared with these state of the art models, as training an equivalent model required significantly larger datasets and more compute resources than we had available.)

Diffusion \citep{sohl-dickstein_deep_2015} machine learning models were developed by applying concepts from non-equilibrium thermodynamics to machine learning, and work by learning the backwards process of removing added noise to attain a generated output. Diffusion models have shown excellent results in generative application in continuous domains, specifically image generation \citep{ho_denoising_2020}.

The continuous nature of denoising in diffusion has posed challenges for applying diffusion to discrete domains such as token generation.  Several approaches have been developed for applying diffusion to discrete domains.  D3PM (2021) \citep{austin_structured_2023} approached diffusion to continuous domains by applying a random probability of corrupting a token to another token during the noising phase, and learning to correct those corruptions during denoising, conditioned on an input.  DiffuSeq (2022) \citep{gong_diffuseq_2023} functions by embedding the discrete tokens into a continuous space, noising and learning to denoise the continuous space, and extracting generated tokens from the denoised continuous space. Score entropy discrete diffusion (SEDD) (2024) \citep{lou_discrete_2024} works by corrupting discrete data with noise, and de-noises using a novel loss function based on the ratio of probabilities of the data between time steps.

\section{Methodology}

% \subsection{LLM acknowledgment}
% The ideas implemented in the project are our own, based on our independent research into the field of diffusion in discrete spaces applied to symbolic regression.  The written report portion of this project was completed without the use of generative AI.  The code used to attain the results was written using a combination of open-source code and datasets, manually written code, and generative AI written code.  GitHub Copilot, Claude 3.7 Sonnet \citep{noauthor_claude_2025}, ChatGPT (4o, O1, O3) \citep{noauthor_chatgpt_2025}, and Gemini 2.5 Pro \citep{noauthor_gemini_2025} were used in various parts of the coding to assist with model implementation and debugging.

% \subsection{Compute resources}
% A workstation equipped with an Intel i7 14700k, 128GB RAM, and 1x Nvidia RTX 3090 with 24GB of VRAM were used for all experiments in this research.

\subsection{Dataset}
As a relatively limited computational resources were available in this research, we tried to ensure that the proposed method and the baseline are compared in the same exact setting. Therefore, a bivariate (2 independent variable) dataset was used for training and evaluating the models below. The dataset from SymbolicGPT \citep{valipour_symbolicgpt_2021} was used for training and evaluating both models.  A total of 500,000 samples were used, split 90/5/5 between train/test/validate.  Each sample consisted of 200 coordinate points.  Samples which contained infinity or NAN values were discarded. 

\subsubsection{Tokenization}
All equations generated were converted to postfix (Reverse Polish Notation) \citep{finkelstein_generalized_2024} before being tokenized. Operators and variables were assigned unique tokens and, and all constants were replaced with a "C" token.  In doing so, the model does not predict constant values, but just the location of constants, which are fit after using conventional algorithms.  This is in contrast to end-to-end symbolic regression models \citep{kamienny_end--end_2022}, which attain state-of-the-art performance by predicting initial constant values before performing conventional fitting for improved performance. End-to-end models require significantly more compute resources to train, so due to limited available resources, our model was not trained to predict constant values. Instead we created a fair setting for comparison, and we compared our proposed method with the baselines without relying on incremental improvements such as beam-search, better constant optimizations by multi-candidate generation, and other minor improvements.

\subsection{Backbone: neural network architecture}

The objective of this work was to obtain a fair comparison between autoregressive generation (Symbolic GPT \citep{valipour_symbolicgpt_2021} baseline)  and D3PM \citep{austin_structured_2023} discrete state-space diffusion (Symbolic Diffusion).  To achieve this, a common neural network architecture was utilized, with equivalent encoders, transformer blocks, and decoder layers were used.  The output layer dimensions and loss functions varied, of course, to accommodate the differences in the architectures used.  Table~\ref{tab:model-hyperparameters} outlines the common model hyperparameters used between both the SymbolicGPT and Symbolic Diffusion models.

\begin{table}[H]
  \caption{Neural Network Hyperparameters}
  \label{tab:model-hyperparameters}
  \centering
  \begin{tabular}{ll}
    \toprule
    Parameter                & Value \\ 
    \midrule
    Embedding Dimension      & 512   \\
    Number of Heads          & 8     \\
    Number of Layers         & 8     \\
    Feedforward Dimension    & 2048  \\
    Dropout                  & 0.15  \\
    \bottomrule
  \end{tabular}
\end{table}

\subsubsection{Encoder}

\begin{figure}[h]
  \centering
  \resizebox{\textwidth}{!}{
    \begin{tikzpicture}[
        font=\sffamily\footnotesize,
        layer/.style={rounded corners=2pt, draw=black!65, very thick,
                      minimum height=1cm, minimum width=2.0cm, % Consistent box width!
                      align=center, inner sep=2pt},
        conv/.style ={layer, fill=blue!15},
        pool/.style ={layer, fill=green!20},
        fc/.style   ={layer, fill=orange!20},
        io/.style   ={layer, fill=gray!18},
        >={Latex[length=2mm,width=2mm]},
        connect/.style={-latex, thick, black},
        node distance = 1.0cm and 1.2cm % consistent, but you can adjust if needed
      ]

      % ---------------- first row -------------------------------------------
      \node[io]   (input) {Input\\Point Cloud\\\small$(B,N,C)$};
      \node[conv, right=of input] (c1)
            {Conv1d\\\small$C\!\rightarrow\!E$\\BN\,{+}\,ReLU};
      \node[conv, right=of c1]    (c2)
            {Conv1d\\\small$E\!\rightarrow\!2E$\\BN\,{+}\,ReLU};
      \node[conv, right=of c2]    (c3)
            {Conv1d\\\small$2E\!\rightarrow\!4E$\\BN\,{+}\,ReLU};

      % ---------------- second row ------------------------------------------
      \node[pool, below=1.2cm of input] (pool)
            {Global Max‑Pool\\\small over $N$ points};

      \node[fc, right=of pool] (f1)
            {Linear\\\small$4E\!\rightarrow\!2E$\\ReLU};
      \node[fc, right=of f1]   (f2)
            {Linear\\\small$2E\!\rightarrow\!E$};
      \node[io, right=of f2]   (output)
            {Condition\\Embedding\\\small$(B,E)$};

      % ---------------- arrows ----------------------------------------------
      \draw[connect] (input) -- (c1);
      \draw[connect] (c1) -- (c2);
      \draw[connect] (c2) -- (c3);

      % S‑curve skip arrow (vertical‑horizontal‑vertical)
      \coordinate (mid1) at ($(c3.south)+(0,-0.75)$);
      \draw[connect]
          (c3.south) -- (mid1)
          -- ($(pool.north |- mid1)$)
          -- (pool.north);

      \draw[connect] (pool) -- (f1);
      \draw[connect] (f1) -- (f2);
      \draw[connect] (f2) -- (output);

      % ---------------- outer frame drawn on background layer ---------------
      \begin{pgfonlayer}{background}
        \node[draw=black, very thick,
              inner sep=6pt,
              rounded corners=0pt,
              fit=(current bounding box)] {};
      \end{pgfonlayer}

    \end{tikzpicture}
  }
  \caption{Architecture of the \texttt{T-Net Encoder}.  
           $B$—batch size, $N$—number of points, $C$—input feature dimensionality,  
           and $E$—embedding dimension.}
  \label{fig:tnetencoder}
\end{figure}
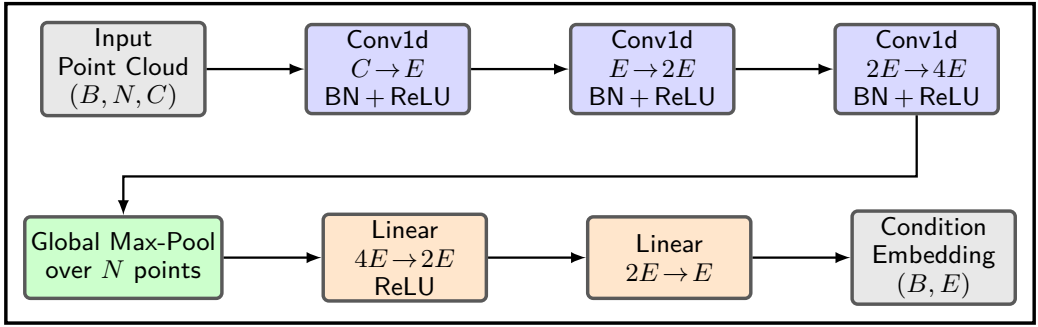

A common T-Net encoder was used, based on the implementation in SymbolicGPT \citep{valipour_symbolicgpt_2021}, which was based on the T-Net developed by Qi et al. \citep{qi_pointnet_2017} for PointNet.  This encoder features 3x blocks of Conv1D, BatchNorm, and ReLU, where the dimension is increased from the input dimension to $E$, $2E$, and $4E$, respectively, where $E$ is the embedding dimension.  Following this, there is a maxpool layer, making the encoder invariant to the number of points in the input feature, followed by 2  linear layers which decrease the dimension from $4E$ to $2E$ and finally the encoder output, of size embedding dimension $E$.
\subsubsection{Transformer blocks and decoder}

As much as possible, identical transformer architectures were used between the Symbolic Diffusion model and the benchmark autoregressive Symbolic GPT model, this common architecture is shown below in Figure \ref{fig:common_transformer_final}.

\begin{figure}[h]
  \centering
  \resizebox{\textwidth}{!}{
    \begin{tikzpicture}[
        font=\sffamily\footnotesize,
        layer/.style={rounded corners=2pt, draw=black!65, very thick,
                      minimum height=1cm, minimum width=1.8cm, % <-- reduced from 2.4cm!
                      align=center, inner sep=2pt},
        conv/.style  ={layer, fill=green!20},
        emb/.style   ={layer, fill=blue!15},
        attn/.style  ={layer, fill=cyan!18},
        ffn/.style   ={layer, fill=orange!20},
        norm/.style  ={layer, fill=yellow!15},
        io/.style    ={layer, fill=gray!18},
        >={Latex[length=2mm,width=2mm]},
        connect/.style={-latex, thick, black},
        node distance = 0.8cm and 1.0cm % <-- reduced for tighter fit
      ]

      % ---------------- point‑cloud path (optional) ---------------------------
      \node[io]   (pc)  {Point Cloud\\\small$(B,N,C)$};
      \node[conv, right=of pc]  (tnet) {T‑Net\\Encoder};
      \node[emb,  right=of tnet] (cond)
            {Condition\\Embedding\\\small$(B,E)$};

      % ---------------- Row 1 : 5 blocks (left‑aligned) -----------------------
      \node[io,   below=0.9cm of pc] (ids)   {Token IDs\\\small$(B,S)$};
      \node[emb,  right=of ids] (tokemb)
            {Token\\Emb.};
      \node[emb,  right=of tokemb] (posemb)
            {+ Pos.\ Enc.};
      \node[emb,  right=of posemb] (merge)
            {Concat Cond.\\+ Proj};
      \node[attn, right=of merge] (attn1)
            {Self‑Attn\\+ Res.};

      % ---------------- Row 2 : 5 blocks (left‑aligned) -----------------------
      \node[ffn,  below=0.9cm of ids] (ffn)
            {FFN\\+ Res.};
      \node[attn, right=of ffn] (stack)
            {$\times L$\\Blocks};
      \node[norm, right=of stack] (ln)
            {LayerNorm};
      \node[emb,  right=of ln]   (proj)
            {Linear\\$E\!\!\to\!\!\text{V}$};
      \node[io,   right=of proj] (logits)
            {Logits};

      % ---------------- arrows ------------------------------------------------
      \draw[connect] (pc) -- (tnet) -- (cond);
      \draw[connect] (cond.south) -- ++(0,-0.4) -| (merge.north);
      \draw[connect] (ids) -- (tokemb) -- (posemb) -- (merge) -- (attn1);
      \coordinate (mid) at ($(attn1.south)+(0,-0.45)$);
      \draw[connect]
          (attn1.south) -- (mid)
          -- ($(ffn.north |- mid)$)
          -- (ffn.north);
      \draw[connect] (ffn) -- (stack) -- (ln) -- (proj) -- (logits);

      % ---------------- bounding box -----------------------------------------
      \begin{pgfonlayer}{background}
        \node[draw=black, very thick, inner sep=6pt, rounded corners=0pt,
              fit=(current bounding box)] {};
      \end{pgfonlayer}

    \end{tikzpicture}
  }
  \caption{Common transformer architecture used by both Symbolic Diffusion and SymbolicGPT.  $B$—batch size, $N$—number of points, $C$—input feature dimensionality,  
           $E$—embedding dimension, and $S$—number of tokens.}
  \label{fig:common_transformer_final}
\end{figure}
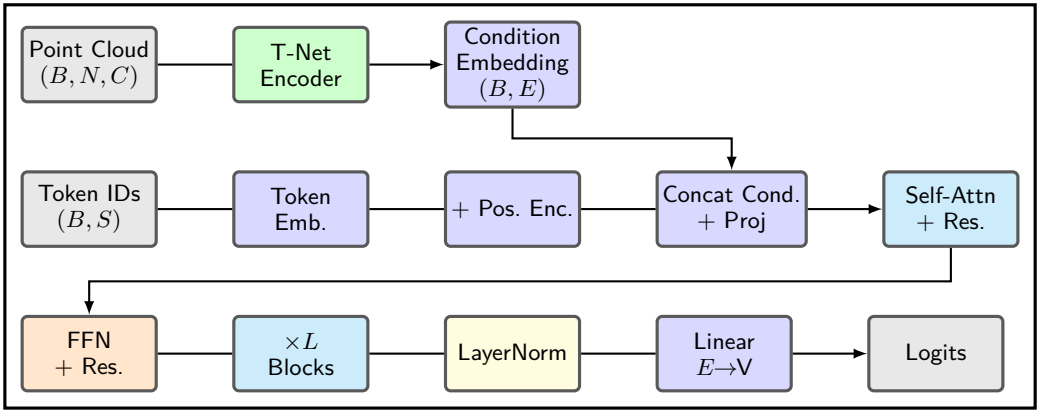

It should be noted that there are some key differences between the architectures due to the different natures of GPT and D3PM diffusion.  Notably, SymbolicGPT uses a causal mask to only attend to prior tokens, where Symbolic Diffusion does not. During inference, Symbolic Diffusion predicts logits for all tokens at timestep $t=0$, whereas SymbolicGPT predicts logits for the next token only. Symbolic Diffusion transformer blocks have global context to all sequences always, whereas SymbolicGPT transformer blocks only have context to already generated tokens.  Diffusion also utilizes a timestep encoding in addition to token position encoding to encode the noising step.  Denoising for D3Pm is completed using the equations discussed below.

\subsection{Proposed method: symbolic regression with discrete state-space diffusion model}
The D3PM model used was based on the paper Structured Denoising Diffusion Models in Discrete State-Spaces by Austin et al. \citep{austin_structured_2023}.  The model was conditioned on the embedded feature vector produced by the encoder.  The hyperparameters from SymbolicGPT \citep{valipour_symbolicgpt_2021} were used as a baseline, and were tuned empirically. A batch size of 64, with a 1e-4 learning rate and AdamW \citep{loshchilov_decoupled_2019} gradient descent were used. Learning rate was reduced by a factor of 0.5 on plateau with a patience of 5 epochs, and early stopping was used with a patience of 15 epochs. For all experiments, a workstation equipped with an Intel i7 14700k, 128GB RAM, and 1x Nvidia RTX 3090 with 24GB of VRAM were used in this research.

\subsubsection{Forward noising process}
D3PM applies diffusion to discrete data, where each entry in x belongs to discrete categories $1,…,K$.  We use a Markovian transition matrix Q to define the probability of any token corrupting to another token with a given probability. 

The forward noising process between timesteps can be represented by the equation, from Austin et al. \citep{austin_structured_2023}:

\begin{equation}
q(x_t \mid x_{t-1}) 
= \mathrm{Cat}\bigl(x_t;\,p = x_{t-1} Q_t\bigr)
\end{equation}

Where x is an array of 1-hot vectors representing the current discrete state of each token, Cat is the categorical distribution, and the multiplication of $x_{(t-1)}$ and $Q_t$ gives the probability of a token’s transition.

The cumulative noising up to time step t is given by, from Austin et al. \citep{austin_structured_2023}:

\begin{equation}
q(x_t \mid x_{0}) 
= \mathrm{Cat}\bigl(x_t;\,p = x_0\,\overline{Q}_t\bigr), 
\quad
\overline{Q}_t = Q_1 Q_2 \cdots Q_t
\end{equation}
A transition matrix with a probability of $(1-\beta_t )$ of remaining at its current token and a $\beta$ probability of randomly being switched to another token was used.  A cosine noising schedule over 1000 noising steps from $\beta= 0.0001$ to $\beta= 0.02$ was used. 

\subsubsection{Reverse noising process}

The denoising process is represented through this equation, provided by Austin et al. \citep{austin_structured_2023},

\begin{equation}
p_{\theta}(x_{t-1}\mid x_{t}) 
\;\propto\;
\sum_{\tilde x_{0}}
q(x_{t-1},x_{t}\mid \tilde x_{0})\,\tilde p_{\theta}(\tilde x_{0}\mid x_{t})
\end{equation}

is the output of the neural network, and we train the neural network to try and predict the probability logits of time step 0 given time step t. this represents the probability of transitioning from $x_{(t-1)}$ to $x_t$, which is known as the forward noising process is defined.  The summation term sums over all possible discrete states for each token being generated, i.e., increasing token length doesn’t increase complexity exponentially. With the proportionality term, we can attain a denoising step from the network’s predictions of the tokens at $x_0$.

An additional important equation provided by Austin et al. \citep{austin_structured_2023} is the posterior, computed through basic application of Baye’s rule,

\begin{equation}
q(x_{t-1}\mid x_t, x_0)
=
\frac{q(x_t\mid x_{t-1},x_0)\,q(x_{t-1}\mid x_0)}{q(x_t\mid x_0)}
=
\mathrm{Cat}\!\Bigl(x_{t-1};\,p = \frac{x_t\,Q_t^{\!\top}\,x_0\,\overline Q_{t-1}}{x_0\,\overline Q_t\,x_t}\Bigr)
\end{equation}

which is used to compute the likelihood of attaining a state at a previous time step given a future time step and knowing the original state. This posterior is needed for the KL divergence contained in the loss function \citep{austin_structured_2023},

\begin{equation}
D_{\mathrm{KL}}\!\bigl[q(x_{t-1}\mid x_t, x_0)\,\|\,p_{\theta}(x_{t-1}\mid x_t)\bigr]
\end{equation}

whereas traditional diffusion uses the variational lower bound as the loss function, a modified loss function was used by the Austin et al. \citep{austin_structured_2023},

\[
L_{\lambda}
\;=\;
L_{\mathrm{vb}}
\;+\;
\lambda\,
\mathbb{E}_{q(x_{0})}\!\Bigl[\,
\mathbb{E}_{q(x_{t}\mid x_{0})}\bigl[-\log\tilde p_{\theta}(x_{0}\mid x_{t})\bigr]
\Bigr].
\]

Where $L_{vb}$ is the standard ELBO variational lower bound used in diffusion using the KL divergence from Sohl-Dickstein et al. \citep{sohl-dickstein_deep_2015}, and the second term represents the cross-entropy for the prediction of $x_0$, which helps improve stability, as the neural network is predicting $x_0$ but it’s prediction is being used for the computation of $x_(t-1)$, so adding cross-entropy with $x_0$ improves the stability of the network’s generation.

\subsection{Inference and constant fitting}
Model inference was done by utilizing the input coordinate points as conditioning vectors to produce tokenized reverse polish notation outputs using both SymbolicGPT and Symbolic Diffusion. 

As the model did not predict constants directly, constant fitting was completed using L-BFGS \citep{liu_limited_1989} implemented from the SciPy library \citep{virtanen_scipy_2020}.  100 iterations of differential evolution \citep{storn_differential_1997} were done to pre-initialize the constants prior to completing L-BFGS fitting, as this was found to produce better outcomes from L-BFGS. This has been done to reduce the error of the L-BFGS which should not affect the comparison of the original models.

\section{Results and discussion}
\subsection{Evaluation metrics}
Two commonly used evaluation metrics for symbolic regression were used, inspired by Kamienny et al. \citep{kamienny_end--end_2022}.  $R^2$ represents how well the predicted model matches the input data \citep{cava_contemporary_2021}, with 1 being a perfect fit.  $R^2$ is unbounded less than 0, so like Kamienny et al. \citep{kamienny_end--end_2022}, the very small number of samples produced with $R^2$ less than 0 were set to zero to prevent excessively bad scores from skewing the mean. Accuracy to tolerance $Acc_\tau$ represents the fraction of equations which the relative error is within a tolerance $\tau$ \citep{biggio_neural_2021, dascoli_deep_2022}.

\begin{equation}
R^2 = 1 - \frac{\sum_{i}^{N_{\text{test}}} (y_i - \hat{y}_i)^2}{\sum_{i}^{N_{\text{test}}} (y_i - \bar{y})^2}, \qquad
\mathrm{Acc}_\tau = \mathbbm{1} \left( \max_{1 \leq i \leq N_{\text{test}}} \left| \frac{\hat{y}_i - y_i}{y_i} \right| \leq \tau \right)
\end{equation}

Additionally, the percentage of valid equations was computed by ensuring the Reverse Polish Notation (RPN) equations were of proper form, checking that no binary operators were attempted on stack sizes of 1, and that the final stack size after all operators is equal to 1.

\subsection{Results}

Both models were evaluated on the same 7000 samples from the validation dataset, presented below in Table \ref{tab:symbolic_comparison}.

%To ensure adequate statistical power for comparing model performance based on $R^{2}$ scores, an a priori power analysis was performed and a target sample size of 6856 pairs was identified to achieve 80\% power ($\alpha$ = 0.05, one-tailed Wilcoxon signed-rank test). Our study analyzed 7000 pairs of samples, and the one-tailed Wilcoxon signed-rank test revealed that the Symbolic Diffusion model's $R^{2}$ scores were statistically significantly higher than those of the SymbolicGPT model (W = 15000, $p$ = 0.01). Since $p < \alpha = 0.05$ we can reject the null hypothesis that Symbolic Diffusion is NOT better than Symbolic GPT and we conclude that there is enough statistically significant evidence to show that the Symbolic Diffusion model's $R^{2}$ scores are higher than the SymbolicGPT model's $R^{2}$ scores. 

\begin{table}[h]
  \caption{Comparison of Symbolic Diffusion and SymbolicGPT on various metrics}
  \label{tab:symbolic_comparison}
  \centering
  \begin{tabular}{lcc}
    \toprule
    Metric & Symbolic Diffusion & SymbolicGPT \\
    \midrule
    Mean $R^2$ Score & \textbf{0.899} & 0.887 \\
    Mean $\mathrm{Acc}_{0.1}$ & 0.655 & \textbf{0.679} \\
    Mean $\mathrm{Acc}_{0.01}$ & 0.531 & \textbf{0.583} \\
    Mean $\mathrm{Acc}_{0.001}$ & 0.425 & \textbf{0.516} \\
    \% Valid RPNs & 0.984 & \textbf{0.999} \\
    \bottomrule
  \end{tabular}
\end{table}

Assuming approximately normal distributions, we can use a paired T-Test to compare outputs.  With the results from the T-Test, we attain that with a 95$\%$ confidence interval, Symbolic Diffusion has a higher mean $R^2$ value (p = 0.001) however, using the same test, SymbolicGPT has a higher (better) mean accuracy than Symbolic Diffusion for all $\tau$ values.  It should be noted that the results between both models are close, and both models show comparable performance overall.  

%As seen in Table \ref{tab:symbolic_comparison}, The mean accuracy scores for all $\tau$ values were better for SymbolicGPT than Symbolic Diffusion. Using the inverse null hypothesis (that Symbolic Diffusion's accuracy tolerance is not better than SymbolicGPT) the Wilcoxon signed-rank test was performed for $\tau = [0.1, 0.01, 0.001]$ and it was found that for all $\tau$, SymbolicGPT's accuracy tolerance is not statistically significantly better than Symbolic Diffusion.  $p = [1, 2, 3]$ respectively indicates that we can not reject the null hypothesis.

\subsection{Discussion}
While Symbolic Diffusion did not outperform the autoregressive SymbolicGPT baseline on all evaluation metrics, it provided a successful demonstration of a novel method of symbolic regression using discrete token diffusion.  Importantly, this demonstrates that for symbolic regression, diffusion based simultaneous generation where the generation process has global context to all tokens throughout the generation process can offer reasonable performance competitive with existing autoregressive models. 

\subsubsection{Societal Impacts}

Improving symbolic regression offers positive societal impacts in the form of improving scientific discovery and improving decision making in society by finding trends in complex data.  Risks exist that these models may capture unwanted biases present in training data, and may risk displacing jobs in data science roles. 

\subsection{Limitations}
There are several important limitations with our work which should be noted.  The intent of this work was not to develop a state-of-the-art model, but rather to compare diffusion generation versus autoregressive generation for the domain of symbolic regression.  Compared to state-of-the-art models (SOTA) such as End-To-End Symbolic Regression \citep{kamienny_end--end_2022}, our model does not predict constants.  Our training dataset of $\sim$ 500,000 was significantly smaller than SOTA models (SymFormer, 130 million \citep{vastl_symformer_2022}, E2E SR $\sim$150 million \citep{kamienny_end--end_2022}).  Furthermore, our model was limited to bivariate data, and was not tested on datasets of varying dimensionalities.  Due to these limitations, our model was unable to be benchmarked against the SOTA conventional and ML based models using tools such as SRBench \citep{cava_contemporary_2021}.  Despite these limitations, our research provides the first fair, apples-to-apples comparison between a diffusion based and an autoregressive based symbolic regression model utilizing nearly identical underlying architectures, providing valuable insights into the performance of previously unexplored diffusion models in the domain of symbolic regression.

We also acknowledge the limited compute that it was available in this research paper and as such we focused on a fair comparison, and the novelty to solve a fundamental issue in the current state-of-the-art transformer models for Symbolic Regression.

\subsection{Future work}
The results demonstrate that diffusion based models are viable and can offer comparable performance to autoregressive models in the domain of symbolic regression. For future work, expanding the dataset size for training and increasing the dimensionality of the training set will be required to evaluate the models performance on more complex data.  Adding end-to-end training with constant prediction will require more compute resources and larger datasets, but may help diffusion based symbolic regression match or exceed state of the art models. Additionally, investigating alternative diffusion models such as masked diffusion \citep{shi_simplified_nodate} may offer improved performance. 

\section{Conclusion}

Our work presents a novel application of discrete-state diffusion modeling to the problem of symbolic regression. By leveraging a D3PM \citep{austin_structured_2023} approach to generate all tokens of an expression in parallel, our model is able to capture global syntactic structures more effectively than left to right autoregressive generation. Through experiments on the bivariate dataset originally designed for SymbolicGPT \citep{valipour_symbolicgpt_2021}, we have shown that our diffusion based framework achieves comparable expression accuracy when using equivalent encoder, transformer, and decoder architecture as a SymbolicGPT \citep{valipour_symbolicgpt_2021} based autoregressive model. 

These results  validate the efficacy of diffusion based generation in the symbolic regression domain, and while further work must still be done to match and surpass autoregressive models, it highlights its potential to mitigate error in tasks requiring syntactical accuracy. 

Overall, our findings demonstrate that discrete diffusion offers a compelling alternative to autoregressive methods for neural network based symbolic regression. We anticipate that further refinement of diffusion models and the incorporation of more complex encodings and architectures will continue to enhance the performance and robustness, ultimately broadening the applicability of neural-network techniques.

\newpage
{
\small

\bibliographystyle{unsrtnat}
\bibliography{references}

\begin{thebibliography}{27}
\providecommand{\natexlab}[1]{#1}
\providecommand{\url}[1]{\texttt{#1}}
\expandafter\ifx\csname urlstyle\endcsname\relax
  \providecommand{\doi}[1]{doi: #1}\else
  \providecommand{\doi}{doi: \begingroup \urlstyle{rm}\Url}\fi

\bibitem[Radwan et~al.()Radwan, Kronberger, and Winkler]{radwan_comparison_2024}
Yousef Radwan, Gabriel Kronberger, and Stephan Winkler.
\newblock \emph{A Comparison of Recent Algorithms for Symbolic Regression to Genetic Programming}.
\newblock \doi{10.48550/arXiv.2406.03585}.

\bibitem[Koza()]{koza_genetic_1994}
John~R. Koza.
\newblock Genetic programming as a means for programming computers by natural selection.
\newblock 4\penalty0 (2):\penalty0 87--112.
\newblock ISSN 1573-1375.
\newblock \doi{10.1007/BF00175355}.
\newblock URL \url{https://doi.org/10.1007/BF00175355}.

\bibitem[Cava et~al.()Cava, Orzechowski, Burlacu, França, Virgolin, Jin, Kommenda, and Moore]{cava_contemporary_2021}
William~La Cava, Patryk Orzechowski, Bogdan Burlacu, Fabrício Olivetti~de França, Marco Virgolin, Ying Jin, Michael Kommenda, and Jason~H. Moore.
\newblock Contemporary symbolic regression methods and their relative performance.
\newblock URL \url{http://arxiv.org/abs/2107.14351}.

\bibitem[Kamienny et~al.()Kamienny, d'Ascoli, Lample, and Charton]{kamienny_end--end_2022}
Pierre-Alexandre Kamienny, Stéphane d'Ascoli, Guillaume Lample, and François Charton.
\newblock End-to-end symbolic regression with transformers.
\newblock URL \url{http://arxiv.org/abs/2204.10532}.

\bibitem[Valipour et~al.()Valipour, You, Panju, and Ghodsi]{valipour_symbolicgpt_2021}
Mojtaba Valipour, Bowen You, Maysum Panju, and Ali Ghodsi.
\newblock {SymbolicGPT}: A generative transformer model for symbolic regression.
\newblock URL \url{http://arxiv.org/abs/2106.14131}.

\bibitem[Vastl et~al.()Vastl, Kulhánek, Kubalík, Derner, and Babuška]{vastl_symformer_2022}
Martin Vastl, Jonáš Kulhánek, Jiří Kubalík, Erik Derner, and Robert Babuška.
\newblock {SymFormer}: End-to-end symbolic regression using transformer-based architecture.
\newblock URL \url{http://arxiv.org/abs/2205.15764}.

\bibitem[Sohl-Dickstein et~al.()Sohl-Dickstein, Weiss, Maheswaranathan, and Ganguli]{sohl-dickstein_deep_2015}
Jascha Sohl-Dickstein, Eric~A. Weiss, Niru Maheswaranathan, and Surya Ganguli.
\newblock Deep unsupervised learning using nonequilibrium thermodynamics.
\newblock URL \url{http://arxiv.org/abs/1503.03585}.

\bibitem[Rombach et~al.()Rombach, Blattmann, Lorenz, Esser, and Ommer]{rombach_high-resolution_2022}
Robin Rombach, Andreas Blattmann, Dominik Lorenz, Patrick Esser, and Björn Ommer.
\newblock High-resolution image synthesis with latent diffusion models.
\newblock URL \url{http://arxiv.org/abs/2112.10752}.

\bibitem[Austin et~al.()Austin, Johnson, Ho, Tarlow, and Berg]{austin_structured_2023}
Jacob Austin, Daniel~D. Johnson, Jonathan Ho, Daniel Tarlow, and Rianne van~den Berg.
\newblock Structured denoising diffusion models in discrete state-spaces.
\newblock URL \url{http://arxiv.org/abs/2107.03006}.

\bibitem[Lou et~al.()Lou, Meng, and Ermon]{lou_discrete_2024}
Aaron Lou, Chenlin Meng, and Stefano Ermon.
\newblock Discrete diffusion modeling by estimating the ratios of the data distribution.
\newblock URL \url{http://arxiv.org/abs/2310.16834}.

\bibitem[Gong et~al.()Gong, Li, Feng, Wu, and Kong]{gong_diffuseq_2023}
Shansan Gong, Mukai Li, Jiangtao Feng, Zhiyong Wu, and Lingpeng Kong.
\newblock {DiffuSeq}: Sequence to sequence text generation with diffusion models.
\newblock URL \url{http://arxiv.org/abs/2210.08933}.

\bibitem[Plasser et~al.()Plasser, Peter, and Widmer]{plasser_discrete_2023}
Matthias Plasser, Silvan Peter, and Gerhard Widmer.
\newblock Discrete diffusion probabilistic models for symbolic music generation.
\newblock pages 5842--5850.
\newblock \doi{10.24963/ijcai.2023/648}.

\bibitem[Li et~al.()Li, Ni, Beardall, Xia, Das, Stan, and Zhao]{li_discdiff_2024}
Zehui Li, Yuhao Ni, William A.~V. Beardall, Guoxuan Xia, Akashaditya Das, Guy-Bart Stan, and Yiren Zhao.
\newblock {DiscDiff}: Latent diffusion model for {DNA} sequence generation.
\newblock URL \url{http://arxiv.org/abs/2402.06079}.

\bibitem[Qi et~al.()Qi, Su, Mo, and Guibas]{qi_pointnet_2017}
Charles~R. Qi, Hao Su, Kaichun Mo, and Leonidas~J. Guibas.
\newblock {PointNet}: Deep learning on point sets for 3d classification and segmentation.
\newblock URL \url{http://arxiv.org/abs/1612.00593}.

\bibitem[noa()]{noauthor_httpscdnopenaicomresearch-coverslanguage-unsupervisedlanguage_understanding_paperpdf_nodate}
https://cdn.openai.com/research-covers/language-unsupervised/language\_understanding\_paper.pdf.
\newblock URL \url{https://cdn.openai.com/research-covers/language-unsupervised/language_understanding_paper.pdf}.

\bibitem[Burlacu et~al.()Burlacu, Yang, and Affenzeller]{burlacu_population_2024}
Bogdan Burlacu, Kaifeng Yang, and Michael Affenzeller.
\newblock Population diversity and inheritance in genetic programming for symbolic regression.
\newblock 23\penalty0 (3):\penalty0 531--566.
\newblock ISSN 1572-9796.
\newblock \doi{10.1007/s11047-022-09934-x}.
\newblock URL \url{https://doi.org/10.1007/s11047-022-09934-x}.

\bibitem[Radford et~al.()Radford, Narasimhan, Salimans, and Sutskever]{radford_improving_nodate}
Alec Radford, Karthik Narasimhan, Tim Salimans, and Ilya Sutskever.
\newblock Improving language understanding by generative pre-training.

\bibitem[Vaswani et~al.()Vaswani, Shazeer, Parmar, Uszkoreit, Jones, Gomez, Kaiser, and Polosukhin]{vaswani_attention_2023}
Ashish Vaswani, Noam Shazeer, Niki Parmar, Jakob Uszkoreit, Llion Jones, Aidan~N. Gomez, Lukasz Kaiser, and Illia Polosukhin.
\newblock Attention is all you need.
\newblock URL \url{http://arxiv.org/abs/1706.03762}.

\bibitem[Ho et~al.()Ho, Jain, and Abbeel]{ho_denoising_2020}
Jonathan Ho, Ajay Jain, and Pieter Abbeel.
\newblock Denoising diffusion probabilistic models.
\newblock URL \url{http://arxiv.org/abs/2006.11239}.

\bibitem[Finkelstein()]{finkelstein_generalized_2024}
Edward Finkelstein.
\newblock Generalized fixed-depth prefix and postfix symbolic regression grammars.
\newblock URL \url{http://arxiv.org/abs/2410.08137}.

\bibitem[Loshchilov and Hutter()]{loshchilov_decoupled_2019}
Ilya Loshchilov and Frank Hutter.
\newblock Decoupled weight decay regularization.
\newblock URL \url{http://arxiv.org/abs/1711.05101}.

\bibitem[Liu and Nocedal()]{liu_limited_1989}
Dong~C. Liu and Jorge Nocedal.
\newblock On the limited memory {BFGS} method for large scale optimization.
\newblock 45\penalty0 (1):\penalty0 503--528.
\newblock ISSN 1436-4646.
\newblock \doi{10.1007/BF01589116}.
\newblock URL \url{https://doi.org/10.1007/BF01589116}.

\bibitem[Virtanen et~al.()Virtanen, Gommers, Oliphant, Haberland, Reddy, Cournapeau, Burovski, Peterson, Weckesser, Bright, van~der Walt, Brett, Wilson, Millman, Mayorov, Nelson, Jones, Kern, Larson, Carey, Polat, Feng, Moore, VanderPlas, Laxalde, Perktold, Cimrman, Henriksen, Quintero, Harris, Archibald, Ribeiro, Pedregosa, van Mulbregt, and {SciPy 1.0 Contributors}]{virtanen_scipy_2020}
Pauli Virtanen, Ralf Gommers, Travis~E. Oliphant, Matt Haberland, Tyler Reddy, David Cournapeau, Evgeni Burovski, Pearu Peterson, Warren Weckesser, Jonathan Bright, Stéfan~J. van~der Walt, Matthew Brett, Joshua Wilson, K.~Jarrod Millman, Nikolay Mayorov, Andrew R.~J. Nelson, Eric Jones, Robert Kern, Eric Larson, C~J Carey, İlhan Polat, Yu~Feng, Eric~W. Moore, Jake VanderPlas, Denis Laxalde, Josef Perktold, Robert Cimrman, Ian Henriksen, E.~A. Quintero, Charles~R. Harris, Anne~M. Archibald, Antônio~H. Ribeiro, Fabian Pedregosa, Paul van Mulbregt, and {SciPy 1.0 Contributors}.
\newblock {SciPy} 1.0: Fundamental algorithms for scientific computing in python.
\newblock 17:\penalty0 261--272.
\newblock \doi{10.1038/s41592-019-0686-2}.

\bibitem[Storn and Price()]{storn_differential_1997}
Rainer Storn and Kenneth Price.
\newblock Differential evolution – a simple and efficient heuristic for global optimization over continuous spaces.
\newblock 11\penalty0 (4):\penalty0 341--359.
\newblock ISSN 1573-2916.
\newblock \doi{10.1023/A:1008202821328}.
\newblock URL \url{https://doi.org/10.1023/A:1008202821328}.

\bibitem[Biggio et~al.()Biggio, Bendinelli, Neitz, Lucchi, and Parascandolo]{biggio_neural_2021}
Luca Biggio, Tommaso Bendinelli, Alexander Neitz, Aurelien Lucchi, and Giambattista Parascandolo.
\newblock Neural symbolic regression that scales.
\newblock URL \url{http://arxiv.org/abs/2106.06427}.

\bibitem[d'Ascoli et~al.()d'Ascoli, Kamienny, Lample, and Charton]{dascoli_deep_2022}
Stéphane d'Ascoli, Pierre-Alexandre Kamienny, Guillaume Lample, and François Charton.
\newblock Deep symbolic regression for recurrent sequences.
\newblock URL \url{http://arxiv.org/abs/2201.04600}.

\bibitem[Shi et~al.()Shi, Han, Wang, Doucet, and Titsias]{shi_simplified_nodate}
Jiaxin Shi, Kehang Han, Zhe Wang, Arnaud Doucet, and Michalis~K Titsias.
\newblock Simplified and generalized masked diffusion for discrete data.

\end{thebibliography}

\end{document}